\documentclass{bmvc2k}

\usepackage{enumitem}
\usepackage{array}
\usepackage{balance}
\usepackage{color}
\usepackage[toc,page]{appendix}
\usepackage{bbm}
\usepackage{algorithm}
\usepackage[noend]{algpseudocode} 
\usepackage{bbding}
\usepackage{booktabs}
\usepackage[normalem]{ulem}
\usepackage{setspace}
\usepackage{stfloats}
\usepackage{bm}

\usepackage{amsmath}
\usepackage{amsfonts}       
\usepackage{comment}
\usepackage{wrapfig}

\definecolor{color}{rgb}{0.7, 0., 0.1}

\newcolumntype{C}[1]{>{\centering\let\newline\\\arraybackslash\hspace{0pt}}m{#1}}

\addauthor{Amin Banitalebi-Dehkordi \\Yong Zhang}{{amin.banitalebi,yong.zhang3}@huawei.com}{1}

\addinstitution{
 Huawei Technologies Canada Co., Ltd.\\
 Vancouver, Canada
}

\runninghead{Banitalebi-Dehkordi and Zhang}{Repaint}

\begin{document}

\title{Repaint: Improving the Generalization of Down-Stream Visual Tasks by Generating Multiple Instances of Training Examples}

\maketitle

\begin{spacing}{0.95}


\begin{figure*}[ht]
    \vspace{-10pt}
    \includegraphics[width=1.0\linewidth]{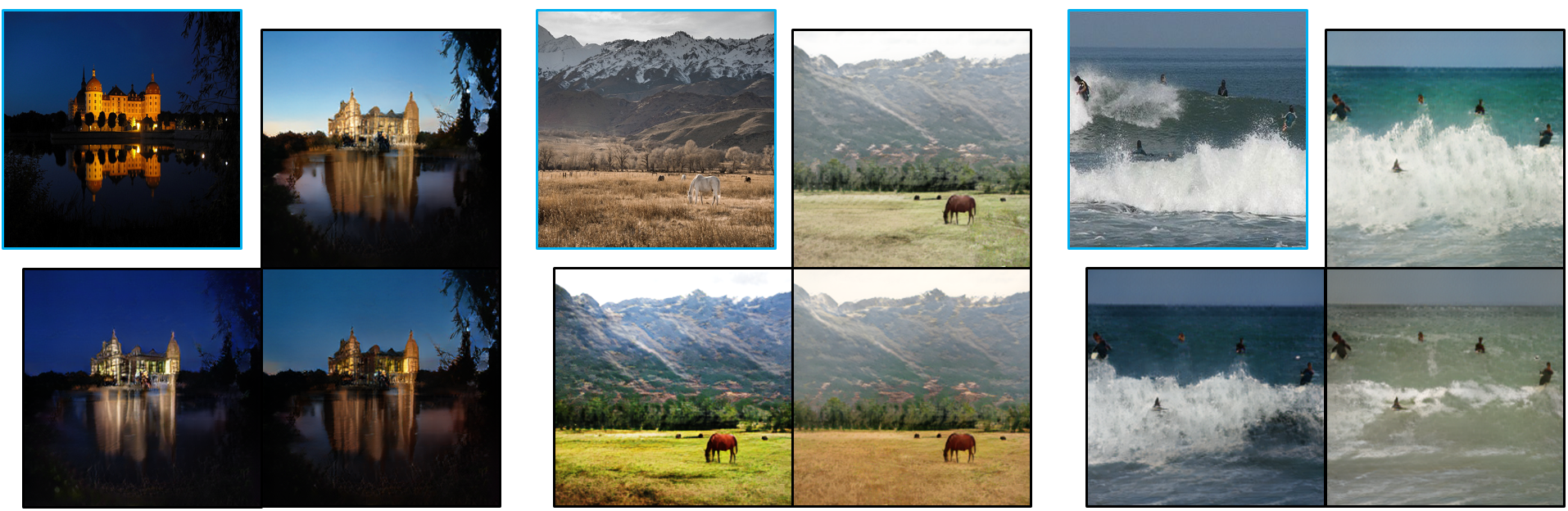}
    \vspace{-18pt}
    \caption{The proposed method augments a training dataset by generating (repainting) an arbitrary number of similar instances to the training examples. Shapes are preserved but texture is diversified. In each case, the top-left corner image is the original example, and the other three demonstrate repainted versions.}
    \label{fig:main-page}
\end{figure*}

\begin{abstract} 
\label{sec:abstract}
Convolutional Neural Networks (CNNs) for visual tasks are believed to learn both the low-level textures and high-level object attributes, throughout the network depth. This paper further investigates the `texture bias' in CNNs. To this end, we regenerate multiple instances of training examples from each original image, through a process we call `\textit{repainting}'. The repainted examples preserve the shape and structure of the regions and objects within the scenes, but diversify their texture and color. Our method can regenerate a same image at different daylight, season, or weather conditions, can have colorization or de-colorization effects, or even bring back some texture information from blacked-out areas. The in-place repaint allows us to further use these repainted examples for improving the generalization of CNNs. Through an extensive set of experiments, we demonstrate the usefulness of the repainted examples in training, for the tasks of image classification (ImageNet) and object detection (COCO), over several state-of-the-art network architectures at different capacities, and across different data availability regimes. Code is released as supplementary \cite{supplementary}.
\end{abstract}

\section{Introduction} \label{sec:introduction}
\vspace{-6pt}
Overfitting is a fundamental problem in training deep neural networks (DNNs) \cite{overfitting}. To overcome the overfitting, there has been a tremendous amount of research which led to many successful approaches including: data augmentation, designing efficient architectures, regularization, dropouts \cite{dropout,dropblock}, early stopping, ensembling, etc. Collectively, these techniques have resulted in achieving a remarkable performance across many applications. That being said, the underlying cause of this problem has also been of interest for a long time. In this paper, we look at this phenomenon through the lens of texture and shape bias \cite{stylize}.


In the case of visual tasks such as object recognition or detection, a common intuition is that deep models such as DCNNs (Deep Convolutional Neural Networks) learn both low-level image features such as edges or texture patterns (within the earlier layers) as well as high-level attributes such as presence and shape of objects (in the deeper layers) \cite{Kriegeskorte, lecun2015deep, stylize}. Some works argue (and sometimes provide empirical results) that like in humans, shape is the single most important factor in CNNs for learning visual tasks \cite{kubilius2016deep, ritter2017cognitive}. Others argue otherwise, that texture has a more significant role in CNNs \cite{gatys2017texture, brendel2019approximating, ballester2016performance, gatys2015texture}. Authors in \cite{stylize} designed a comprehensive study (texture-shape conflict stimuli) to understand this phenomenon. They concluded that CNNs are generally biased towards easier-to-learn texture features (shortcuts) at the expense of shape attributes (texture bias). \cite{stylize} further proposed Shape-ResNet, in which they trained a ResNet model with stylized images and were subsequently able to improve the generalization and robustness of the network. 

Other than the style transfer method used in \cite{stylize}, there has been a family of style transfer algorithms employed for different applications \cite{jing2020dynamic, yim2020filter, na2020multimodal, davis2020text}. However, these approaches generally produce artistic effects on images and diverge from natural-looking images. Moreover, the stylized transfer used in \cite{stylize} is not trained to minimize a down-stream task loss, but rather is an off-the-shelf one \cite{AdaIN}. 

In this paper, we propose a method to augment the training set, by generating multiple instances from each training example. To this end, we make use of a generative semantic synthesis model to generate new instances (in a variational manner), and tie this model to a down-stream task. In other words, we generate examples that adhere to the objects shapes of the original image while modifying the texture in a way that helps the down-stream task (e.g. image classification or object detection). Therefore, our method `repaints' the original images, by changing their texture/color but preserving the shapes and locations of objects. In-place repaint makes it suitable for non-classification down-stream tasks such as object detection. Figure \ref{fig:main-page} demonstrates example images resulted from our method. We verify our approach with an extensive set of experiments for the tasks of image classification and object detection, on several network architectures and dataset sizes.

The main contributions of this paper can be summarized as:
\vspace{-5pt}
\begin{itemize}
    \item We propose a method of augmenting training datasets by repainting the examples. Repainted examples are diverse in texture and color in that they substitute regions, objects, or backgrounds with randomly drawn new instances learned from the dataset. Sometimes this results in interesting outcomes such as adding/removing colors, uncovering new information in blacked-out areas, or shifting day/night time or seasons.
    \vspace{-15pt}
    \item We utilize the repainted training examples to improve the generalization of CNNs. Due to the nature of this method, it can be applied to various visual tasks. We demonstrate results on image classification and object detection as two common use-cases.
    \vspace{-6pt}
    \item We present a comprehensive set of experiments over several state-of-the-art network architectures at different capacities, and across different data availability regimes. Results show a consistent improvement in the generalization of the CNNs.
\end{itemize}



\vspace{-16pt}
\section{Related works} \label{sec:related_works} 
\vspace{-6pt}
In this section, we review the related areas to our work, and draw connections between them.

\vspace{-8pt}
\paragraph{Image generation:}
A number of works such as \cite{bowles2018gan} explore directly augmenting the training data using generative adversarial networks (GANs). These methods train an off-the-shelf GAN with the available images, and later use it to generate more. In this setting, no strong supervision between the GAN and the down-stream task is enforced, and these methods are more useful for situations like medical imaging tasks where data itself is scarce. Other works including \cite{zhu2018emotion} propose class-aware conditioned GANs to balance a dataset for an improved generalization. Our method conditions the generation to be shape preserving while also assisting a general-purpose down-stream task.

\vspace{-8pt}
\paragraph{Semantic image synthesis:}
These methods generate synthetic images given semantic segmentation masks \cite{spade,gaugan}. The idea is to design a GAN (Generative Adversarial Network \cite{gan}) to generate images that can adhere to semantics. In our method, we make use of the spatially adaptive normalization \cite{spade} in order to preserve the shape structure of objects. 

\vspace{-10pt}
\paragraph{Stylization:}
It was argued in \cite{stylize} that ImageNet CNNs are biased towards texture features since they are easier to learn than shape attributes. This shortcut then resulted in a less accurate generalization to unseen test data. To address this issue, authors in \cite{stylize} proposed Shape-ResNet, in which multiple stylized versions of each training image are generated and used for training a CNN (e.g. ResNet). The model is then fine-tuned with the original train set. Since this method is based on style transfer, it preserves the shapes and structures but applies texture modification according to another image's style. In a way, it is also increasing the train set size by augmenting it with the stylized examples. Our findings are in agreement with the observations of \cite{stylize} in that reducing the texture bias can improve the generalization. However, since the stylized examples generated for Shape-ResNet do not look like typical natural images found in standard datasets such as ImageNet \cite{imagenet} or Microsoft COCO \cite{coco}, this can reduce the potential gains (as we see in Section \ref{sec:experiments}). In addition, the style transfer step is detached from the down-stream task (image classification in case of \cite{stylize}), and thus provides no guarantee that the stylized examples can confidently boost the generalization. Nonetheless, there are a variety of style transfer methods proposed in the literature \cite{jing2020dynamic, yim2020filter, na2020multimodal, davis2020text} which may be used similarly.

\vspace{-10pt}
\paragraph{Learning to modify input examples:}
Related to our approach is a line work where the input training images are updated according to some loss term that is related to a down-stream task \cite{resize, calibration, zhu2018data}. For example, authors in \cite{resize} propose to learn to resize input data, in such a way that can help with down-stream tasks of image classification or quality assessment. Our method is in some sense similar since we also learn to update the input data, however, it is different in the sense that we keep the original image size but instead learn to regenerate and replace (repaint) objects in the image. Therefore, the two methods are orthogonal and can be combined with each other.

\vspace{-10pt}
\paragraph{Image augmentation:}
Image augmentation has a rich literature. Traditionally, global image-scale operations such as rotate, flip, blur, contrast stretch, etc. were used within augmentation pipelines. Over the past several years, many new augmentation techniques were proposed. These techniques include MixUp \cite{mixup}, CutOut \cite{cutout}, CutMix \cite{cutmix}, AutoAugment \cite{autoaugment}, Thumbnail \cite{thumbnail}, ClassMix \cite{classmix}, etc. Our method is orthogonal to these kinds of image augmentations and in fact these augmentations can be applied on top of our method. We provide some results in this regard in Section \ref{sec:experiments}.

\vspace{-6pt}
\section{The proposed repainting method} 
\label{sec:method}

\vspace{-4pt}
In this section we first introduce some basic setup, then explain our solution followed by several remarks and discussions.

\begin{figure*}
    \centering
    \vspace{-2pt}
    \includegraphics[width=1.0\linewidth]{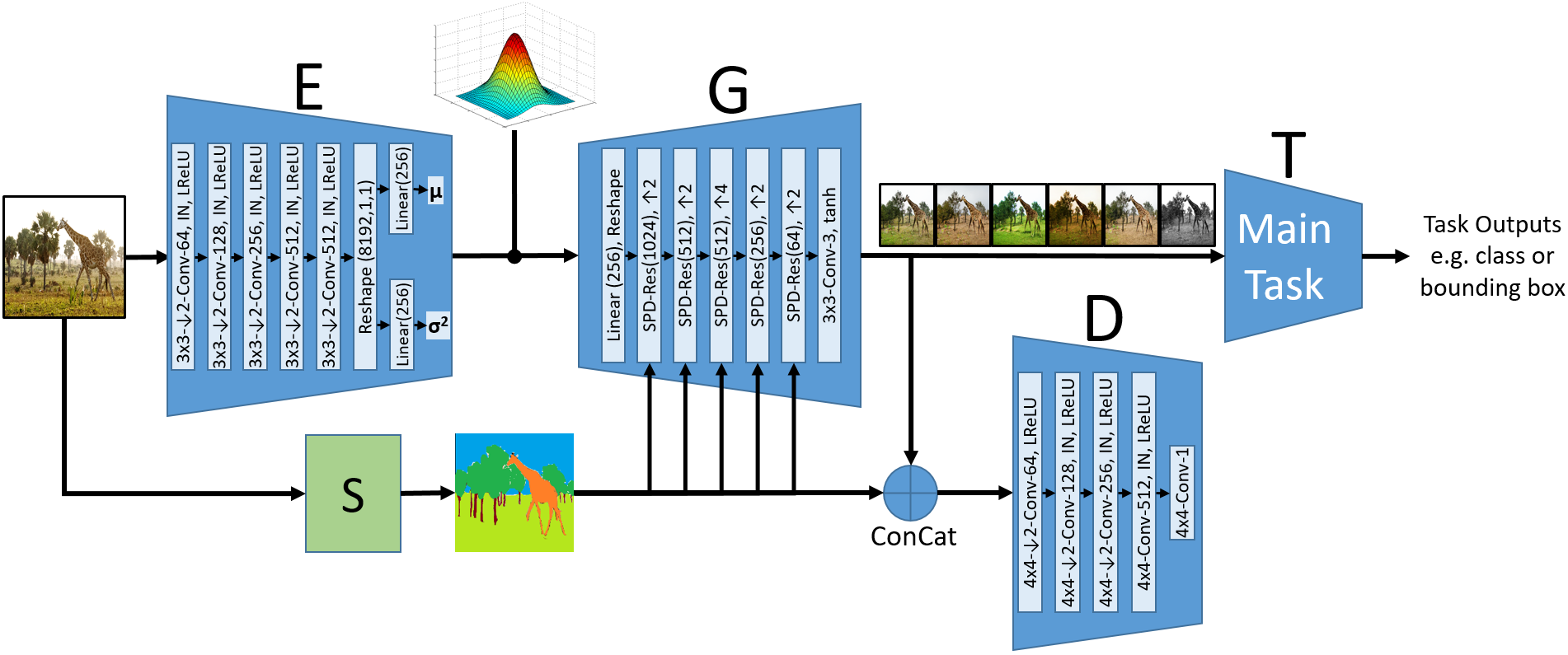}
    \vspace{-22pt}
    \caption{The proposed method employs a VAE-GAN-like architecture \cite{larsen2016autoencoding} that is tied to the down-stream task. E, G, D, and T denote the encoder, generator, discriminator, and the main task. S is a module that generates approximate segmentation masks, such as a DeepLab model \cite{deeplab}, or the image-based Felzenszwalb-Huttenlocher (FH) \cite{FH} algorithm. There is no restriction on what the down-stream task can be, except that its input is an image. We examine image classification and object detection tasks in this paper.}
    \label{fig:flow-diagram}
\end{figure*}

\vspace{-8pt}
\subsection{Some basic setup}
\vspace{-4pt}
Let $E$, $D$, $G$, and $T$ denote the encoder, discriminator, generator, and down-stream task in our setup, respectively. Also, let $S$ denote a module that generates some form of semantic segmentation mask (details in the next subsection). We incorporate the aforementioned modules in our design, as shown in Figure \ref{fig:flow-diagram}. 

Moreover, we make use of the SPatially Adaptive DEnormalization (SPADE) modules introduced in \cite{spade}. The SPADE module and its corresponding residual block denoted by SPD-Res enforce the consistency of shapes and structures, and can be formulated as:
\vspace{-4pt}
\begin{equation}
    \label{eq:spade}
    {f_{out}}^i_c = \gamma^i_c \frac{{f_{in}}^i_c-\mu^i_c}{\sigma^i_c} + \beta^i_c,
    \vspace{-4pt}
\end{equation}
where $f_{in}$ and $f_{out}$ are the input and output feature maps of shape $B\times{N_c}\times{H}\times{W}$, $B$ is the mini-batch size, $N_c$ is the number of channels, $H$ and $W$ denote the height and width of the activations tensor in (\ref{eq:spade}), $\mu^i_c$ and $\sigma^i_c$ are mean and standard deviations of input features, $i$ and $c$ denote the layer and channel indices, and $\gamma$ and $\beta$ are learned scale and bias modulation tensors (with spatial dimensions) that are multiplied and added element-wise to output of a sync-BN layer to create the output features. It can be observed in (\ref{eq:spade}) that this kind of normalization is in some ways similar to regular batch normalization, but it has spatial dimensions that are learned, which in turn helps with enforcing shapes and structures. 

\begin{wrapfigure}{r}{0.54\linewidth}
    \vspace{-10pt}
    \centering
    \includegraphics[width=0.95\linewidth]{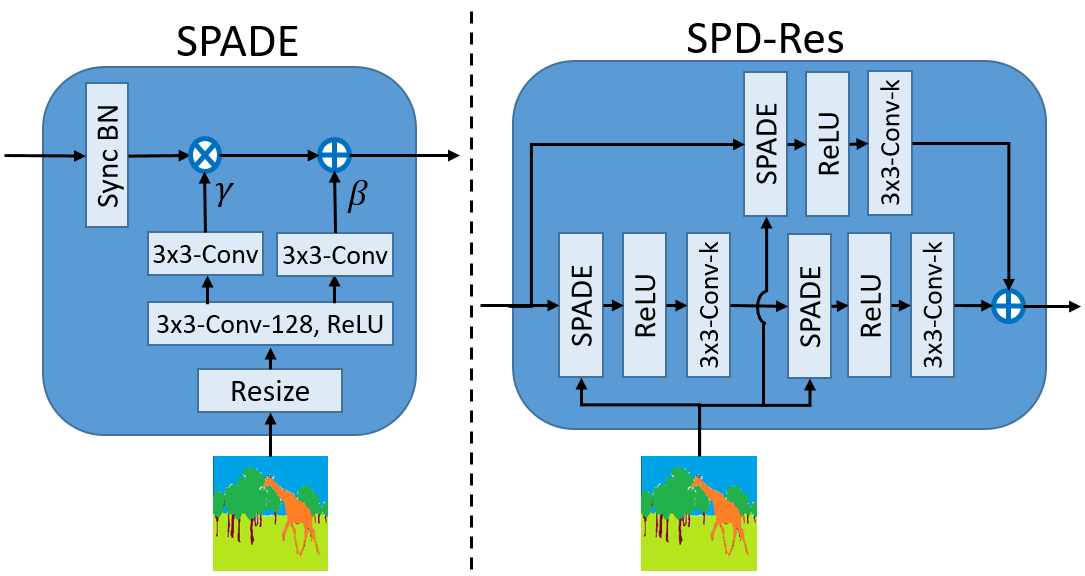}
    \vspace{-8pt}
    \caption{SPADE and its residual block SPD-Res.}
    \label{fig:spade}
\end{wrapfigure}

Figure \ref{fig:spade} illustrates the inner architecture of SPADE and its residual block. Note that $\gamma$ and $\beta$ in (\ref{eq:spade}) and Figure \ref{fig:spade}, for each layer, are of shape $N_c\times{H}\times{W}$. Unlike the standard BatchNorm scale and bias parameters they depend on the spatial mask values. In some sense, (\ref{eq:spade}) is like applying segmentations to the activation maps, thereby conditioning the shapes and structures.


\subsection{Architecture design and loss}
Figure \ref{fig:flow-diagram} shows the flow-diagram of our method. As observed from this figure, there is a similarity (in terms of the overall architecture) to traditional image generation or semantic image synthesis architectures. The encoder and generator together generate batches of repainted images. The discriminator is responsible for pushing the generated examples to look `real'. Module $S$ outputs segmentation masks from the input images. Note that the segmentation masks from this module don't necessarily have to be very accurate. In our experiments in Section \ref{sec:experiments}, we provide results based on using masks generated by a DeepLab-v2 \cite{deeplab} model, as well as masks generated by a completely unsupervised image-based operator of \cite{FH}, and show that in both cases we can achieve generalization gains (the visualizations of repainted images are based on using the DeepLab-v2 masks). Finally, the $T$ module refers to the task network. For a classification task, it could be any classification CNN e.g. ResNet \cite{resnet} or EfficientNet \cite{efficientnet} with a softmax layer at the end. Or, for an object detection task, it could be any detection network such as a YOLO \cite{yolo} or EfficientDet \cite{efficientdet} model. During training, each image is seen once per epoch, but is repainted slightly differently every time.

The training objective and loss terms follow those of \cite{spade} (and thus also pix2pixHD \cite{pix2pixHD}), however we add a new loss term for training the down-stream task. The overall objective therefore contains three terms: $\mathcal{L}_{Generator-Encoder}$ to account for the encoder and generator losses, $\mathcal{L}_{Discriminator}$ to compute the discriminator loss, and finally $\mathcal{L}_{Task}$ to denote the task loss. The $\mathcal{L}_{Generator-Encoder}$ itself accounts for the generator loss ($\mathcal{L}_{Generator}$), feature matching loss ($\mathcal{L}_{Feat.}$ used in \cite{pix2pixHD}), and a KLD loss to account for the variational sampling of the encoder's output ($\mathcal{L}_{KLD}$). (\ref{eq:loss}) and (\ref{eq:loss-g}) summarize the above. We refer the readers to \cite{pix2pixHD} for further details on the generator and discriminator loss terms.
\vspace{-4pt}

\begin{equation}
    \label{eq:loss}
    \mathcal{L} = \mathcal{L}_{G} + \mathcal{L}_{D} + \mathcal{L}_{Task} = \mathcal{L}_{GAN} + \mathcal{L}_{Feat.} + \mathcal{L}_{KLD} + \mathcal{L}_{Task}.
\end{equation}
\vspace{-10pt}
\begin{equation}
    \label{eq:loss-g}
    \mathcal{L}_{GAN}(G,D) = \mathbb{E}[logD(.)] + \mathbb{E}[log(1 - D(G(.)))].
\end{equation}

The task loss $\mathcal{L}_{Task}$ accounts for task-specific losses. For classification task it will be a cross-entropy loss, and for object detection it will be a detection loss according to specialized detection architectures (usually a regression loss to account for bounding boxes, a cross-entropy loss for objects category assignment, and a confidence score loss).

Training is done iteratively similar to GANs \cite{gan}, however, we train in iterations for the terms in (\ref{eq:loss}).
To this end, on one iteration the model is run through, real images are run through D, and then D and T are updated using the discriminator loss and task loss. In the other iteration the model is run through, and E and G are updated based on the generator loss and task loss. As a result, the task loss would also be supervising E and G.
We observed in our experiments that the best results were achieved when optimizing two iterations of discriminator and down-stream task, and one iteration for the generator. More details are given in Section \ref{sec:experiments}.

\begin{figure*}
    \includegraphics[width=1.0\linewidth]{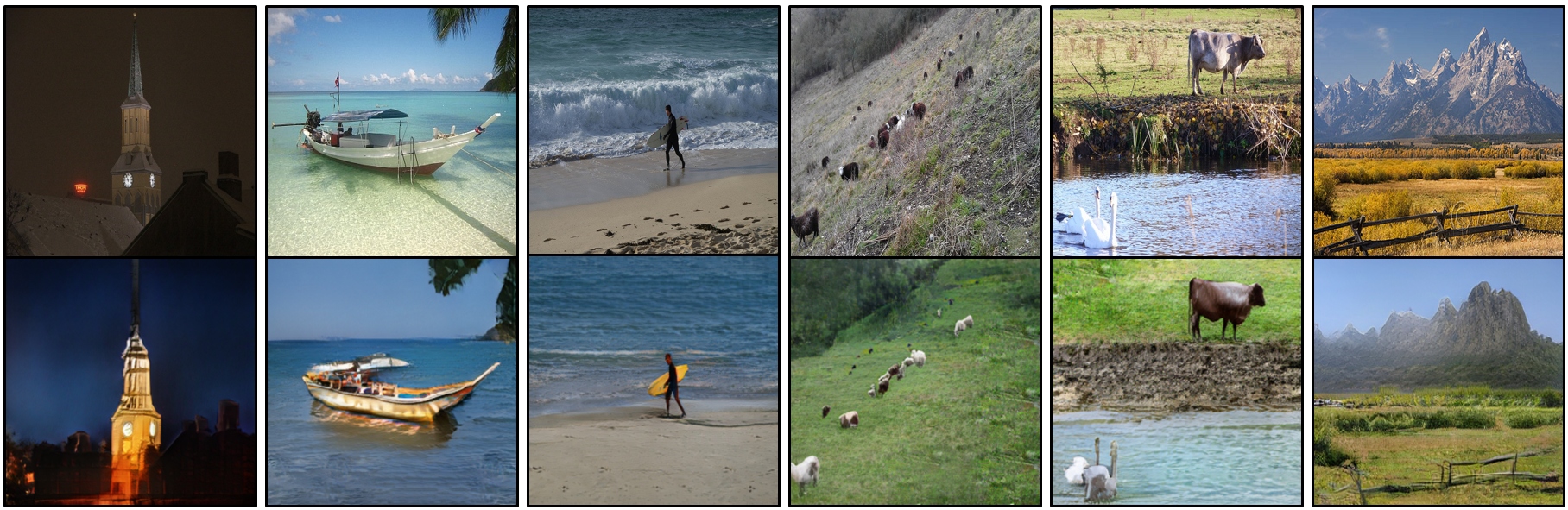}
    \vspace{-20pt}
    \caption{Repaint examples. The top row shows the original samples, and the bottom row shows the repainted versions. Notice the texture changes such as: tower lights, boat reflection and design, beach waves pattern, animal bodies or land coverage, and mountain snow.}
    \label{fig:examples-pairs-short}
\end{figure*}

\vspace{-4pt}
\subsection{Remarks and discussions} \label{sec:remarks}
\vspace{-4pt}
Here we discuss some remarks about our approach.

\vspace{-10pt}
\paragraph{Consistent shapes, diversified texture:}
The variational sampling from the encoder's latent space and the SPADE blocks together result in images generated with a similar structure and objects shapes to the input, but with a new texture and color style that is randomly sampled from what the the model has learned from the texture and colors of previous examples. It is therefore like replacing/repainting each object/region with a new instance learned from the same distribution. Due to the texture bias phenomenon \cite{stylize}, this can improve the network's generalization. Figure \ref{fig:main-page} and \ref{fig:examples-pairs-short} show examples of the generator's output.

\begin{figure*}
\centering
    \includegraphics[width=0.85\linewidth]{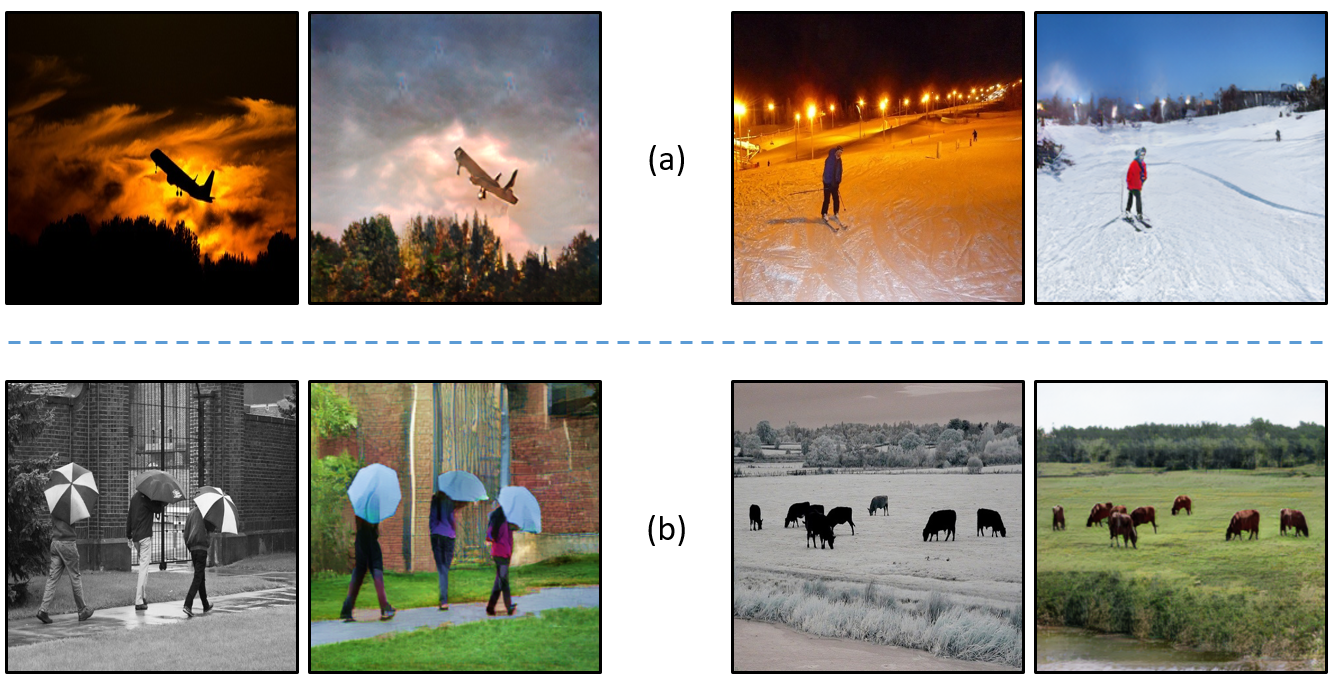}
    \vspace{-8pt}
    \caption{Examples of repainted images where new information is uncovered. a: uncovering from the dark things such as trees texture, mountain trees, or a building afar. b: colorization. For each pair, the left-side image is the original, and the right-side one is a repaint.}
    \label{fig:examples-uncover}
\end{figure*}

\vspace{-10pt}
\paragraph{Colorization, time shift, uncovering new texture, etc.:}
We observe an interesting effect in the repainted images where sometimes they demonstrate effects such as: colorization, de-colorization, changing the time of the day or seasons, or even bringing back textures which were blacked out in the original images. Figure \ref{fig:examples-uncover} demonstrates examples of such effects.

\vspace{-10pt}
\paragraph{Reuse of bounding boxes:}
Since the labels for the down-stream tasks (e.g. classes or bounding boxes) do not change, a repainted image may contain substitute instances of the same object categories present in the scene. The in-place repaint allows to perform tasks such as object detection since the locations of ground truth bounding boxes do not change. Hence, the same set of ground truth labels can be used for the augmented images.

\vspace{-10pt}
\paragraph{Image generation:}
The goal of our method is not necessarily to generate visually pleasant or normal-looking images, but rather is to generate images that are suitable for the down-stream task. That being said, by including a discriminator loss, and carefully balancing the iterative training of the generator, discriminator, and task modules, we achieve an acceptable look on the generated examples.

\vspace{-10pt}
\paragraph{Architecture:}
The general architecture and layers of the E, D, and G blocks are inspired by \cite{pix2pixHD, spade}, but customized for our purpose to be attached to down-stream tasks such as classification or detection. In particular, we have designed these blocks to be somewhat light-weight, as the main task can have a large burden on the GPU memory during training. For example, in Section \ref{sec:experiments} we tested our method with down-stream task of EfficientNet-D3 object detection that has 25B FLOPs (roughly 64$\times$ more operations and 2$\times$ more parameters than EfficientNet-B0 classification model). That being said, the GAN modules can be replaced with different architectures used in the generative models literature, as long as they enforce the shape consistency like we do.

\vspace{-10pt}
\paragraph{Orthogonality with augmentations or regularizations:}
When training the main task, repainting is orthogonal to other kinds of image augmentations, and thus they can be applied at the same time. In fact, we show in Section \ref{sec:experiments} that an improved generalization can be achieved by applying augmentations such as CutMix \cite{cutmix} or regularizations such as DropBlock \cite{dropblock} together with repainting.

\vspace{-10pt}
\paragraph{Limitations:}
An observation we made during our experiments with natural image datasets such as ImageNet and COCO is that our method is very good at repainting scenes in general, however, it sometimes has a hard time with finer details such as facial features. Examples of such failure cases are provided in the supplementary materials \cite{supplementary}. It is also worth noting that specialized GANs such as the ones used for faces, are trained with face datasets, whereas we used general purpose datasets such as ImageNet or COCO that contain a wide range of scenes and objects, and may not be very suitable for specialized tasks. That being said, we expect the image generation to perform well when trained on specialized and controlled datasets. Moreover, the goal is not to necessarily generate good looking images, but it is to generate images that help the down-stream task (reflected in the task performance results). In addition, as mentioned in Section \ref{sec:training-details}, after training with repainted images for a while, at the end we fine-tune with the original dataset. This ensures the task performance will be protected from infrequent shortfalls of image generation.

\vspace{-5pt}
\section{Experiments} \label{sec:experiments}
\vspace{-3pt}
We discuss the experiment results in this section. To this end, we first explain the datasets and metrics used, followed by the training details and baselines. Then we discuss our results and ablation studies.

\vspace{-5pt}
\subsection{Datasets and metrics}
\vspace{-3pt}
Our experiments include two down-stream tasks of image classification and object detection. For classification, we use the ImageNet dataset \cite{imagenet} with 1.28M training and 50K validation examples. The main metric of performance is top-1 or top-5 classification accuracy (\%).

For object detection, we use the Microsoft COCO dataset \cite{coco} with 118K training and 5K validation examples. Methods are assessed based on mean Average Precision (mAP) metric either at a certain IoU (Intersection over Union) threshold such as 0.5, or averaged over various IoUs e.g. @0.5:0.95.

\vspace{-4pt}
\subsection{Training details} \label{sec:training-details}
\vspace{-2pt}
Baselines include training various architecture at different model capacities. For classification, we use MobileNet-v2 \cite{mobilenetv2}, ResNet50 \cite{resnet}, and the EfficinetNet \cite{efficientnet} family of B0, B1, B2, and B3. For object detection, we include EfficientDet-D0, D1, D2, and D3 \cite{efficientdet}. Implementations were in PyTorch v1.6 and included customized code from the following repositories: timm \cite{timm} commit 532e3b4, efficientdet \cite{whitemaneffdet} commit 1c9a3d3, and SPADE \cite{spadecode} commit 1a687ba.

There are a large number of training experiments done for the two tasks of classification and detection which increases diversity in the training procedures, but in general we followed a 200 epoch training strategy. We used the learning rates of $1e^{-4}$ and $4e^{-4}$ for the generator and discriminator, respectively, with an Adam optimizer \cite{adam} with $\beta_1=0$ and $\beta_2=0.999$. In addition, for the down-stream task of classification, we used a learning rate of 0.12, with decays of 90\% every 3 epochs, and a rmsprop optimizer \cite{ruder2016overview} with warm-up. Similarly for the task of object detection, we used a learning rate of 0.06, rmsprop with cosine decay rule, and warmup. Baseline models were trained for 200 epochs. For our method, we first trained for 150 epochs, and then performed a 50 epoch fine-tuning of only the task part of the model with the original dataset. Note that longer training may result in slightly better performance. In fact, some state-of-the-art (SOTA) ImageNet models are trained for 500 epochs. That being said, our comparisons are fair and produce accuracies close to those of SOTA models.

Also note that repainting happens on the fly for each training example. In that sense, all the training images are used in each epoch, and each time a random repaint is applied.

Moreover, each training job was run on a 8-GPU node with V100 GPUs of 32GB memory, and was repeated 5 times to ensure the consistency of the results. 

It is also worth noting that in our experiments we never used any ground truth semantic segmentation masks. We used approximate masks generated by a DeepLab-v2 model \cite{deeplab}, as well as rough masks generated by the Felzenszwalb-Huttenlocher (FH) \cite{FH} method. The FH algorithm is a classical image-based method and is unsupervised in nature.

\subsection{Main results}
Table \ref{tab:results-imagenet} shows the results of image classification experiments. It is observed from Table \ref{tab:results-imagenet} that, by increasing the diversity of examples during training, repainting consistently improves the classification accuracy across several models with different capacities. In Table \ref{tab:results-comparisons-imagenet}, we compare the results of our method with another image-generation based method, Shape-ResNet \cite{stylize}, as well as several augmentation-based and regularization methods. Note that in general our method is orthogonal to augmentation-based or regularization-based approaches, and therefore its performance is expected to improve when combined with such approaches. This is consistent with our observations in Table \ref{tab:results-comparisons-imagenet}.

\begin{table}[!b]
\centering

\fontsize{8}{10}\selectfont
\begin{tabular}[t]{lrrccr}
\toprule
Architecture & \# params & FLOPs & Baseline & Repaint w/o task-loss & Ours (Repaint)\\
\midrule
MobileNetv2 & 3.4M & 0.3B & 74.63 & 74.94 & 75.15 (\textcolor{blue}{+0.52})\\
ResNet-50 & 26M & 4.1B & 77.06 & 77.64 & 78.05 (\textcolor{blue}{+0.99})\\
EffNet-B0 & 5.3M & 0.39B & 76.66 & 77 & 77.24 (\textcolor{blue}{+0.58})\\
EffNet-B1 & 7.8M & 0.70B & 78.59 & 78.92 & 79.15 (\textcolor{blue}{+0.56})\\
EffNet-B2 & 9.2M & 1.0B & 79.24 & 79.61 & 79.83 (\textcolor{blue}{+0.60})\\
EffNet-B3 & 12M & 1.8B & 80.87 & 81.2 & 81.45 (\textcolor{blue}{+0.58})\\
\bottomrule
\end{tabular}
\caption{\label{tab:results-imagenet} \small Top-1 accuracy for image classification on ImageNet.}

\end{table}

\begin{table}[!b]
\centering

\fontsize{8}{10}\selectfont
\begin{tabular}[t]{lrrccr}
\toprule
Architecture & \# params & FLOPs & Baseline & Repaint w/o task-loss & Ours (Repaint)\\
\midrule
EffDet-D0 & 3.9M & 2.5B & 33.87 & 34.36 & 35.10 (\textcolor{blue}{+1.23})\\
EffDet-D1 & 6.6M & 6B & 38.98 & 39.38 & 39.97 (\textcolor{blue}{+0.99})\\
EffDet-D2 & 8.1M & 11B & 42.25 & 42.69 & 43.35 (\textcolor{blue}{+1.10})\\
EffDet-D3 & 12.0M & 25B & 45.27 & 45.98 & 46.86 (\textcolor{blue}{+1.59})\\
\bottomrule
\end{tabular}
\caption{\label{tab:results-coco} \small mAP performance for object detection on COCO.}

\end{table}

It is also worth noting that applying Repaint without the task-specific loss (separate optimizations, i.e. a trained/frozen generator) can still lead to improvements over the baseline as it increases data diversity. However, the incorporation of the task loss can further boost the performance since it also encourages the generation to assist with the down-stream task.

Furthermore, as mentioned in Section \ref{sec:remarks} the proposed method does not rely on a specific type of generative network. Different architectures employed in the generative models' literature can also be used, as long as they enforce the shape consistency. One such example is \cite{spade}. That being said, the original architecture of \cite{spade} is relatively large; when attached to a large task network such as a large detection model, it will consume a large amount of GPU memory. This enforces a very small batch size which makes the training on large datasets such as COCO or ImageNet impractical. For the sake of comparisons however, we added results of \cite{spade} plus task-loss on ResNet50 to Table \ref{tab:results-comparisons-imagenet}. We observe that this benchmark achieves a comparable top-1 accuracy to Repaint, suggesting that the extra generation capacity did not necessarily directly translate to a considerably better down-stream top-1.

Table \ref{tab:results-coco} shows the results of object detection experiments. We observe from Table \ref{tab:results-coco} that object detection training also benefits from repainting by a considerable margin.

In conclusion, the results obtained by our main experiments are inline with the observations of \cite{stylize} in that texture bias has an important role in the training of CNN models, and texture diversification leads to generalization improvements.

\begin{table}
\centering

\fontsize{8}{10}\selectfont
\begin{tabular}[t]{lC{2.4cm}C{1.2cm}}
\toprule
Method & Strategy & Top-1 (\%)\\
\midrule
Baseline &  & 77.06\\
MixUp \cite{mixup} & Augmentation & 77.9\\
CutOut \cite{cutout} & Augmentation & 77.1\\
CutMix \cite{cutmix} & Augmentation & 78.6\\
AutoAugment \cite{autoaugment} & Augmentation & 77.6\\
DropBlock \cite{dropblock} & Regularization & 78.1\\
ISDA \cite{wang2019implicit, wang2021regularizing} & Latent Augmentation & 78.1\\
\midrule
Shape-ResNet \cite{stylize} (rerun) & Image Generation & 77.42\\
\cite{spade} + task-specific loss & Image Generation & 78.02\\
Ours (Repaint) & Image Generation & 78.05\\
\midrule
CutMix+Repaint & Combo & 79.11 \\
CutMix+DropBlock+Repaint & Combo & 79.19 \\
\bottomrule
\end{tabular}
\caption{\label{tab:results-comparisons-imagenet} \small A comparison of ImageNet top-1 classification accuracy on ResNet-50. Repaint combined with other augmentations or regularizations can yield a high performance.}

\end{table}

%

\subsection{Ablation studies}

Next, we perform ablation studies on the proposed method. We first investigate the effect of segmentation mask quality. To this end, we try DeepLab-v2 and FH masks to consider both high and low quality masks. Table \ref{tab:results-imagenet-delta} and Table \ref{tab:results-coco-delta} show the gains over baselines achieved by our method for classification and detection across many models, over 5 runs. Note that in Table \ref{tab:results-imagenet-delta} and \ref{tab:results-coco-delta}, the min-max intervals from the 5 runs are computed by measuring the gap between the best/worst baseline runs and the worst/best repaint runs. We observe from these tables that improvements are consistent, although slightly lower for the FH masks. 

In another ablation study, we investigate the impact of labels availability in lower data regimes. To this end, we report results when 1\% or 10\% of training data is used. As observed in Table \ref{tab:results-portion-classification} and \ref{tab:results-portion-detection}, higher gains are achieved when lower portions of data are used. This is somewhat expected since ImageNet and COCO datasets contain a large number of examples, and achieving better generalization with 100\% of examples is therefore more difficult.

All-in-all, repainting shows a robust and consistent improvement in the generalization capability of image classification and object detection tasks, and therefore can be considered as an add-on orthogonal candidate for inclusion in existing training pipelines.

\begin{table}
\centering

\fontsize{8}{10}\selectfont
\begin{tabular}[t]{lC{2.7cm}C{2.7cm}}
\toprule
Architecture & Gains with DeepLab-v2 (min,max) over 5 runs & Gains with FH \cite{FH} (min,max) over 5 runs\\
\midrule
MobileNetv2 & +0.52 (0.38,0.69) & +0.44 (0.28,0.60)\\
ResNet-50 & +0.99 (0.71,1.28) & +0.79 (0.51,1.03)\\
EffNet-B0 & +0.58 (0.41,0.75) & +0.51 (0.34,0.68)\\
EffNet-B1 & +0.56 (0.39,0.72) & +0.48 (0.31,0.65)\\
EffNet-B2 & +0.60 (0.42,0.77) & +0.50 (0.33,0.67)\\
EffNet-B3 & +0.58 (0.40,0.76) & +0.49 (0.29,0.66)\\
\midrule
Average & \textcolor{blue}{+0.77 (0.54,0.99)} & \textcolor{blue}{+0.64 (0.41,0.85)}\\
\bottomrule
\end{tabular}
\caption{\label{tab:results-imagenet-delta} \small Ablation on magnitude and consistency of performance improvements across weakly-supervised or unsupervised masks; Results on ImageNet top-1 classification accuracy (\%).}

\end{table}

\begin{table}
\centering

\fontsize{8}{10}\selectfont
\begin{tabular}[t]{lC{2.7cm}C{2.7cm}}
\toprule
Architecture & Gains with DeepLab-v2 (min,max) over 5 runs & Gains with FH \cite{FH} (min,max) over 5 runs\\
\midrule
EffDet-D0 & +1.23 (1.01,1.46) & +1.02 (0.83,1.21)\\
EffDet-D1 & +0.99 (0.84,1.14) & +0.78 (0.60,0.96)\\
EffDet-D2 & +1.10 (0.93,1.29) & +0.89 (0.71, 1.07)\\
EffDet-D3 & +1.59 (1.31,1.87) & +1.17 (0.88,1.46)\\
\midrule
Average & \textcolor{blue}{+1.22 (1.17,1.44)} & \textcolor{blue}{+0.96 (0.75,1.18)}\\
\bottomrule
\end{tabular}
\caption{\label{tab:results-coco-delta} \small Ablation on magnitude and consistency of performance improvements across weakly-supervised or unsupervised masks; Results on COCO object detection accuracy (mAP \%).}

\end{table}

\begin{table}[!ht]
\centering

\fontsize{8}{10}\selectfont
\begin{tabular}[t]{lcccccc}
\toprule
Method & \multicolumn{3}{c}{Top-1} & \multicolumn{3}{c}{Top-5}\\ 
\cmidrule(l{3pt}r{3pt}){2-4} \cmidrule(l{3pt}r{3pt}){5-7}
& 1 \% & 10 \% & 100 \% & 1 \% & 10 \% & 100 \% \\
\midrule
Baseline & 14.63 & 57.13 & 77.06 & 31.61 & 79.60 & 93.57\\
Ours (Repaint) & 18.72 & 59.55 & 78.05 & 35.55 & 81.13 & 93.97\\
\midrule
Gains & \textcolor{blue}{+4.09} & \textcolor{blue}{+2.42} & \textcolor{blue}{+0.99} & \textcolor{blue}{+3.94} & \textcolor{blue}{+1.53} & \textcolor{blue}{+0.40}\\
\bottomrule
\end{tabular}
\caption{\label{tab:results-portion-classification} \small ResNet-50 classification accuracy vs portion of ImageNet dataset used for training.}

\end{table}

\begin{table}[!ht]
\centering

\fontsize{8}{10}\selectfont
\begin{tabular}[t]{lcccccc}
\toprule
Method & \multicolumn{3}{c}{mAP @0.50:0.95} & \multicolumn{3}{c}{AP @0.50}\\ 
\cmidrule(l{3pt}r{3pt}){2-4} \cmidrule(l{3pt}r{3pt}){5-7}
& 1 \% & 10 \% & 100 \% & 1 \% & 10 \% & 100 \% \\
\midrule
Baseline & 6.21 & 22.44 & 33.87 & 23.82 & 39.06 & 52.36\\
Ours (Repaint) & 10.35 & 24.73 & 35.10 & 27.95 & 41.58 & 53.71\\
\midrule
Gains & \textcolor{blue}{+4.14} & \textcolor{blue}{+2.29} & \textcolor{blue}{+1.23} & \textcolor{blue}{+4.13} & \textcolor{blue}{+2.52} & \textcolor{blue}{+1.35}\\
\bottomrule
\end{tabular}
\caption{\label{tab:results-portion-detection} \small EfficientDet-D0 object detection accuracy vs portion of COCO dataset used for training.}

\end{table}


\vspace{-16pt}
\paragraph{Computational complexity}
Compared to training only a task network (e.g. a classification or detection network), our method requires an additional generative module to repaint the images. This can be thought of as a learned augmentation, and thus incurs an extra overhead. 
Since the GAN part is designed to be relatively light-weight, and the segmentation piece is not being trained, the overall computational complexity is still comfortably manageable.
On average, the training time of the image classification and object detection tasks observed an increase of $\approx66\%$ and $\approx53\%$, respectively. The overhead can be reduced by further compressing the generative network module, by offline training and freezing it, or by applying the repainting only on a percentage of the training examples.



\section{Conclusion} \label{sec:conclusion}
\vspace{-4pt}
In this paper, we proposed a method of augmenting a training dataset by variationally repainting the training images. The images generated by our method were diverse in texture and color but all preserved the original shape and structure. We then leveraged the augmented dataset to train models with improved generalization on test data. We demonstrated the performance of our method on the tasks of image classification (ImageNet) and object detection (COCO), over several state-of-the-art network architectures at different capacities, and across different data availability regimes. We hope our work can help facilitate further research in this direction.


\end{spacing}

\clearpage
{\small
\bibliography{references}
}

\clearpage
\section{Supplementary materials} \label{sec:supplementary}

This section contains the supplementary materials.

\subsection{Source code}
We share our implementation code to make it easy to reproduce our results. The source-code is attached to the supplementary materials in a `code' directory. We also provide detailed instructions for training and evaluating our models in `README.md' files.




\subsection{Future works} \label{sec:future}

In this paper we mostly focused on the texture bias \cite{stylize}. Future works include studying the shape bias more carefully. We hope to answer the question of `Can we employ a technique of image generation, to diversify the shape attributes in a controlled way that can further help a down-stream task?'. Another direction for future work, is to investigate cases such as fine-grained classification on birds species, and evaluate the impact of texture and shape bias. These kinds of tasks might require special considerations/conditioning when generating new instances.

\subsection{Additional visualizations} 
\begin{figure}
    \centering
    \includegraphics[width=0.7\linewidth]{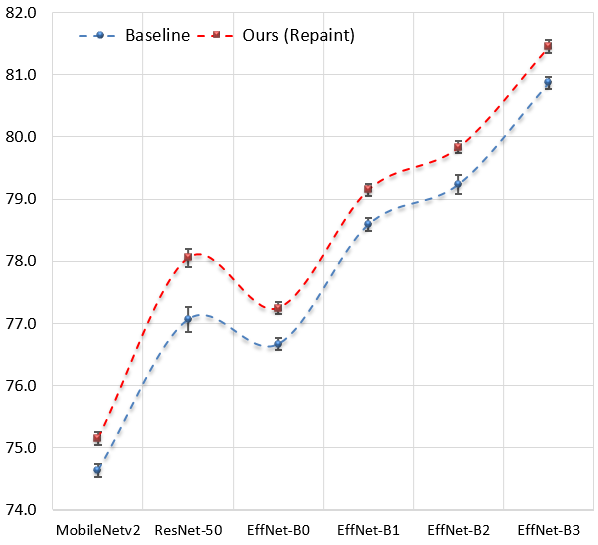}
    \vspace{-8pt}
    \caption{Performance evaluations on the ImageNet dataset. Top-1 (\%) classification accuracy for baselines and repaint. Error-bars demonstrate consistent improvements.}
    \label{fig:results-imagenet}
\end{figure}

\begin{figure}
    \centering
    \includegraphics[width=0.7\linewidth]{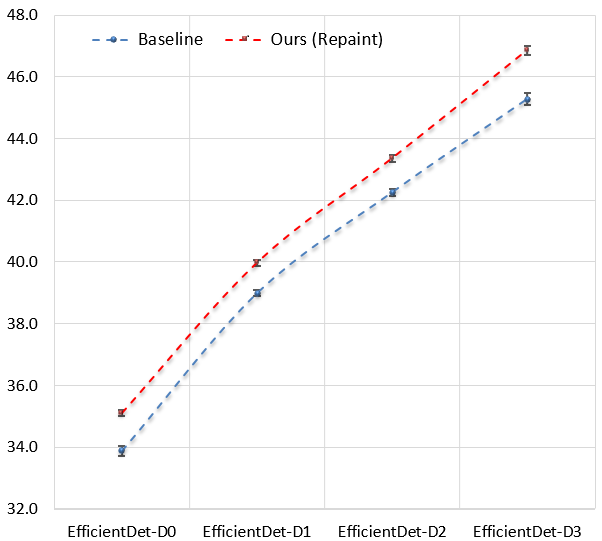}
    \vspace{-8pt}
    \caption{Performance evaluations on the COCO dataset. mAP @0.50:0.95 (\%) for baselines and repaint. Error-bars demonstrate statistically consistent improvements.}
    \label{fig:results-coco}
\end{figure}

Figure \ref{fig:results-imagenet} and \ref{fig:results-coco} show the classification and object detection performance of various models for baselines and our method. The error bars of multiple runs demonstrate the statistical consistency of the improvements.

\begin{figure*}
    \includegraphics[width=1.0\linewidth]{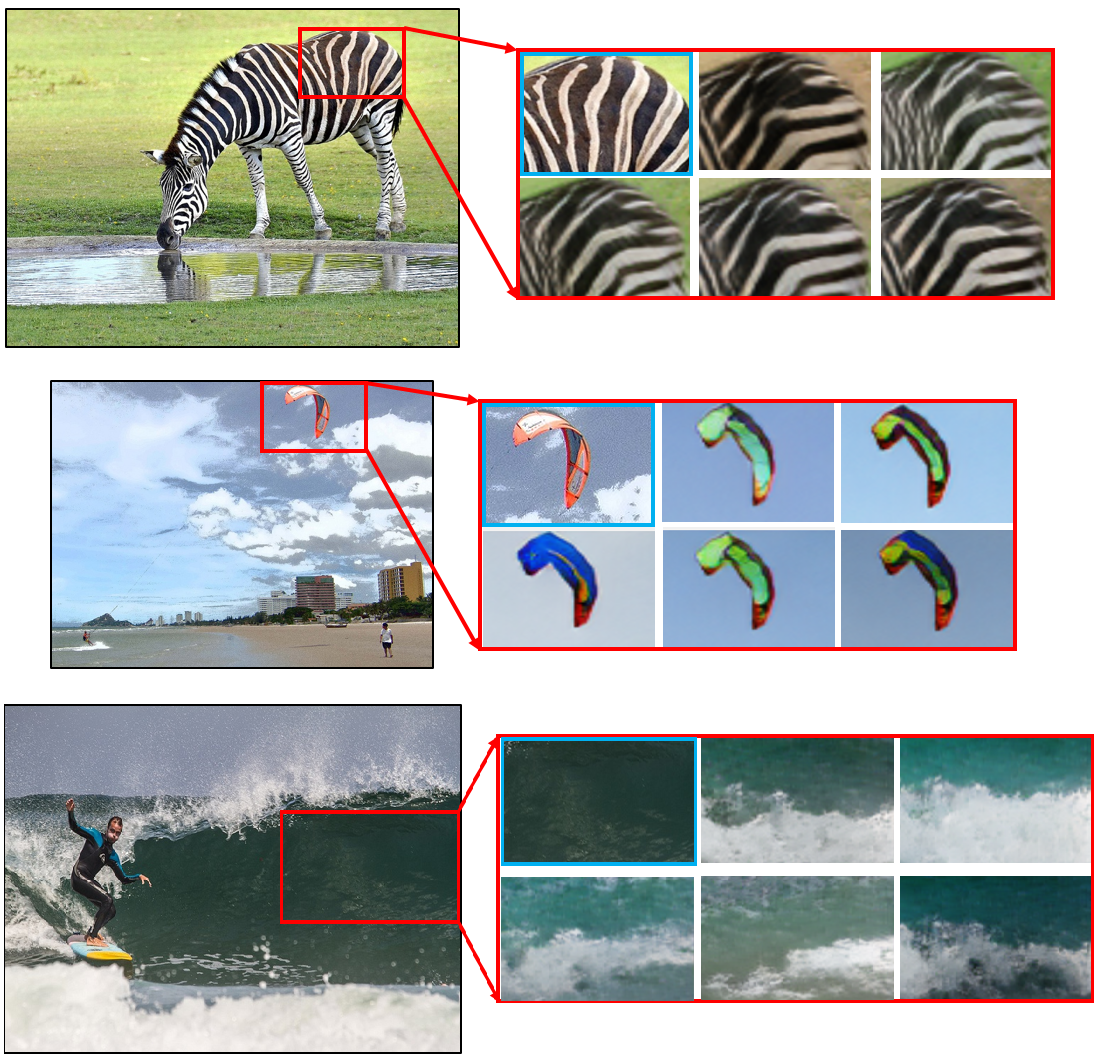}
    \vspace{-8pt}
    \caption{Both texture and color are repainted. Here we compare an original patch (top-left) with multiple repainted instances.}
    \label{fig:examples-zoom}
\end{figure*}

\begin{figure*}
    \includegraphics[width=1.0\linewidth]{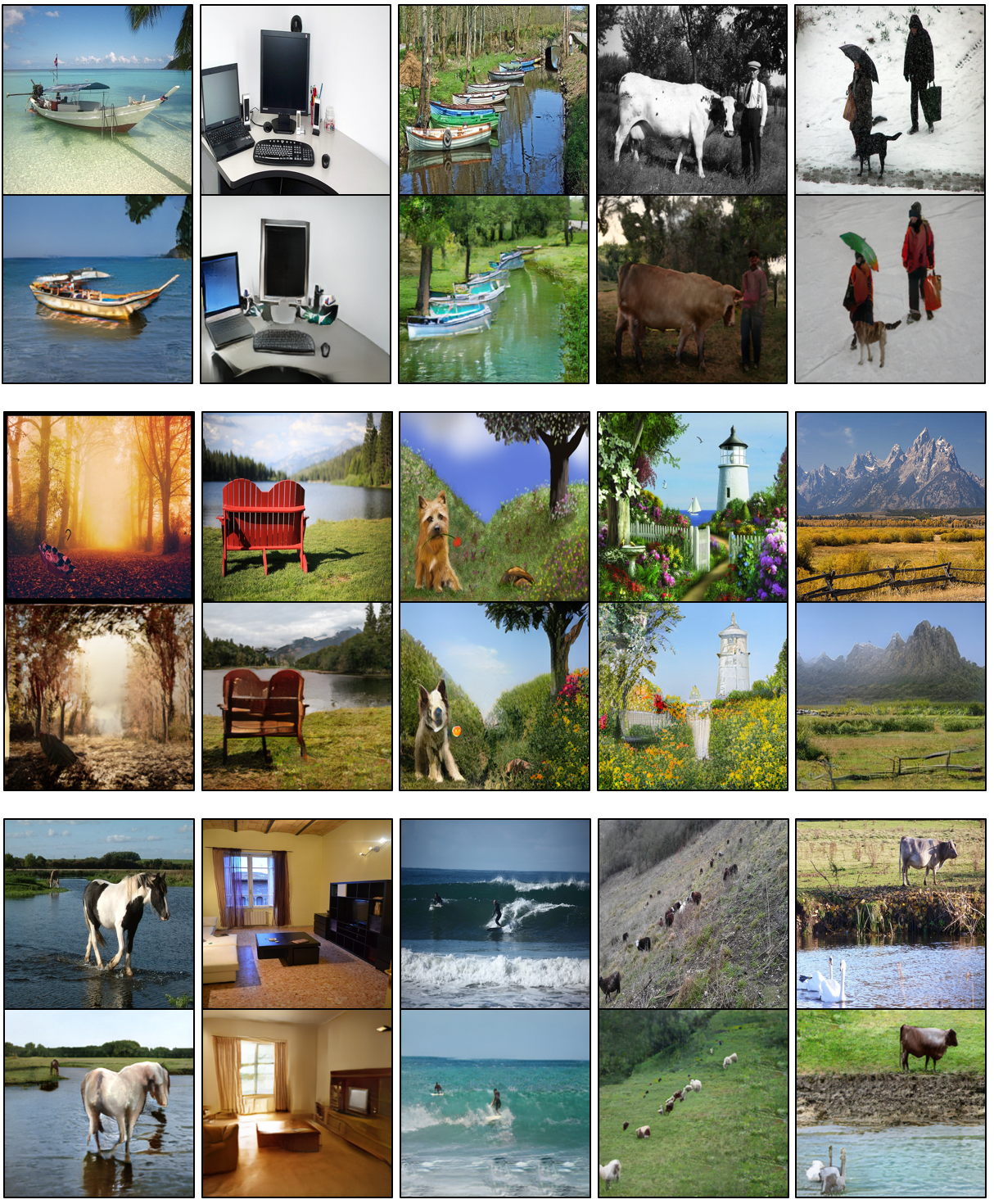}
    \vspace{-8pt}
    \caption{Example generated images in pairs: in each case, the upper row is an original image, and the lower row shows the repainted version.}
    \label{fig:examples-pairs-1}
\end{figure*}

\begin{figure*}
    \includegraphics[width=1.0\linewidth]{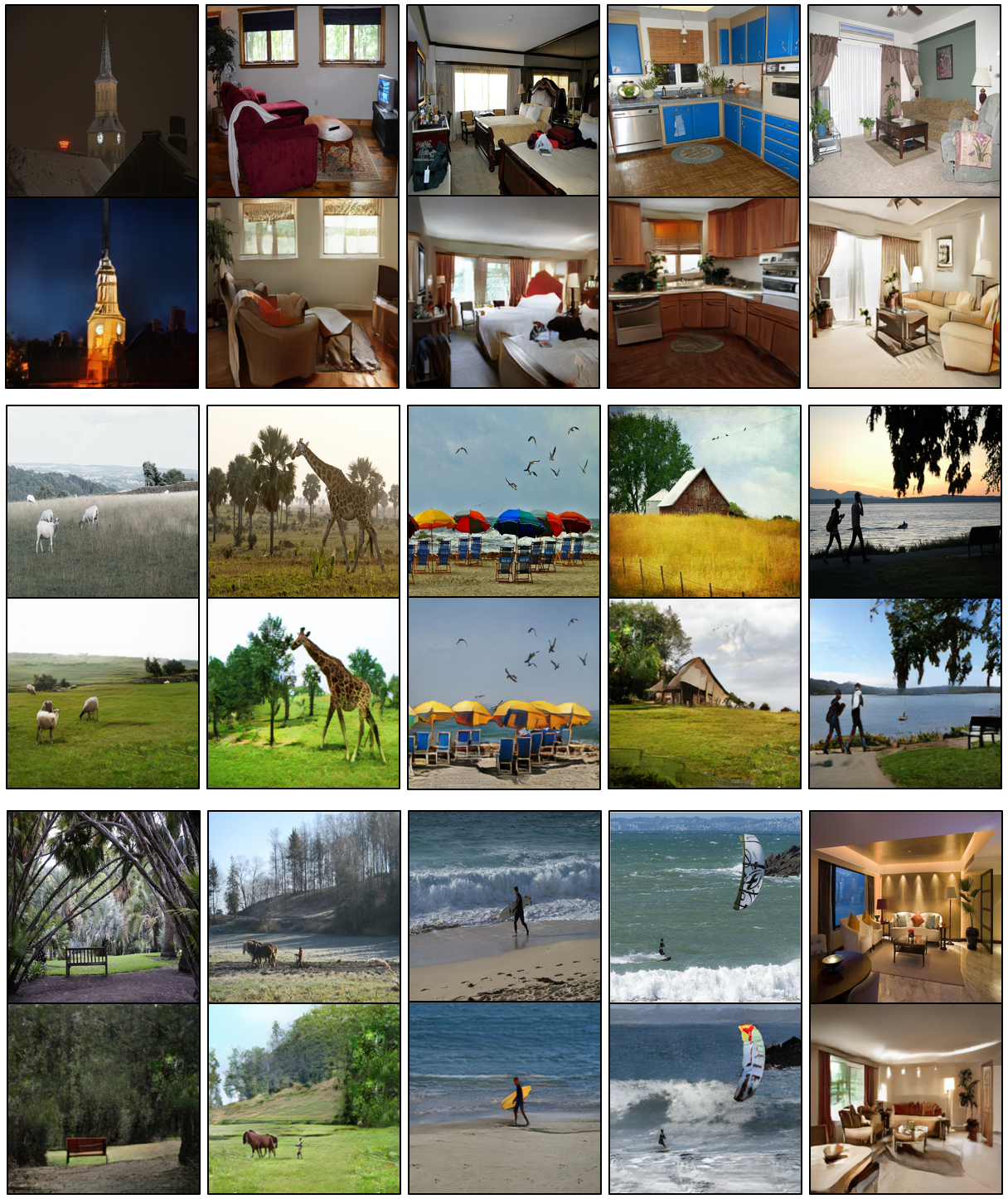}
    \vspace{-8pt}
    \caption{Example generated images in pairs: in each case, the upper row is an original image, and the lower row shows the repainted version.}
    \label{fig:examples-pairs-2}
\end{figure*}

\begin{figure*}
    \includegraphics[width=1.0\linewidth]{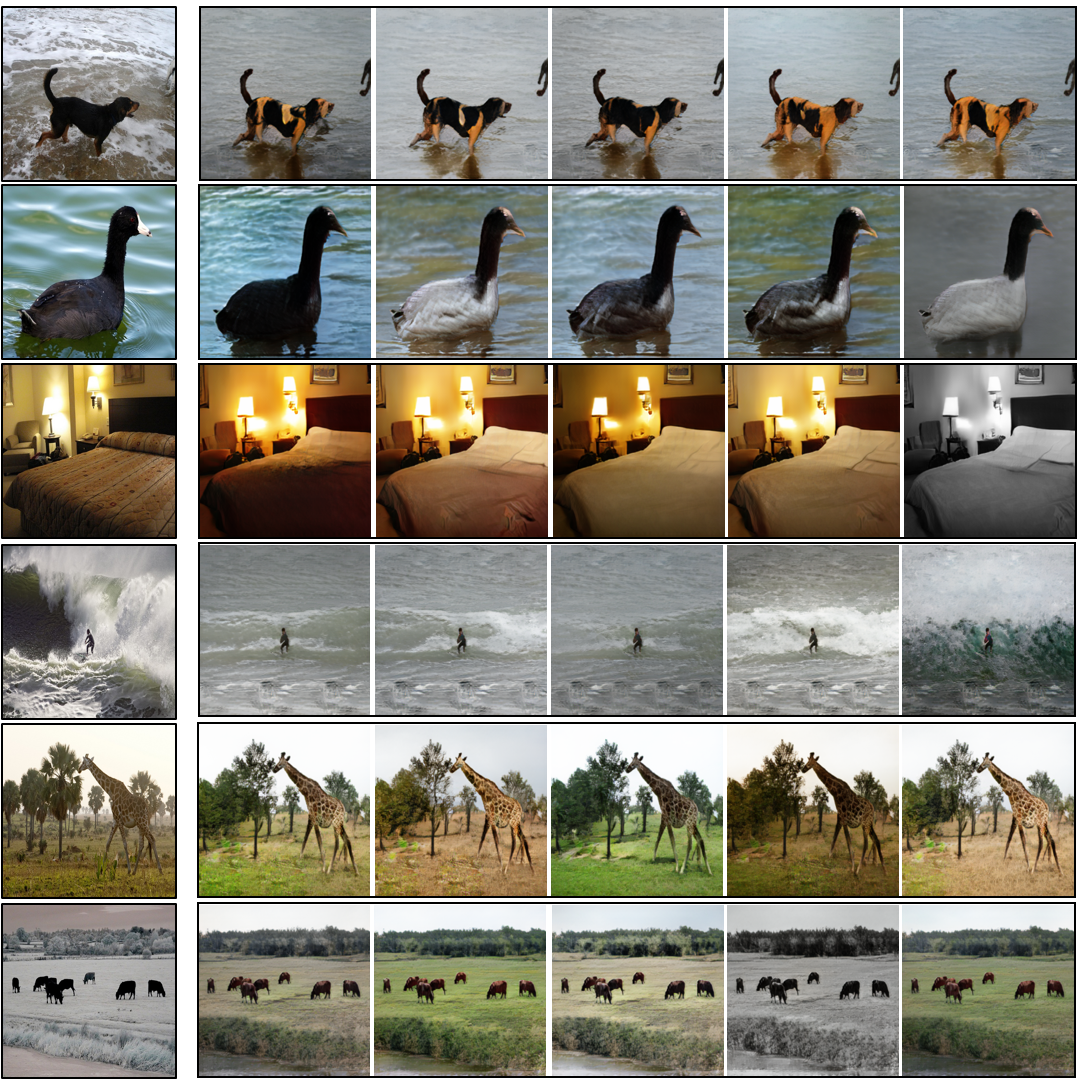}
    \vspace{-8pt}
    \caption{Example generated images from the COCO dataset: in each case, the left column is an original image, and the other columns show the repainted versions.}
    \label{fig:examples-coco-1}
\end{figure*}

\begin{figure*}
    \includegraphics[width=1.0\linewidth]{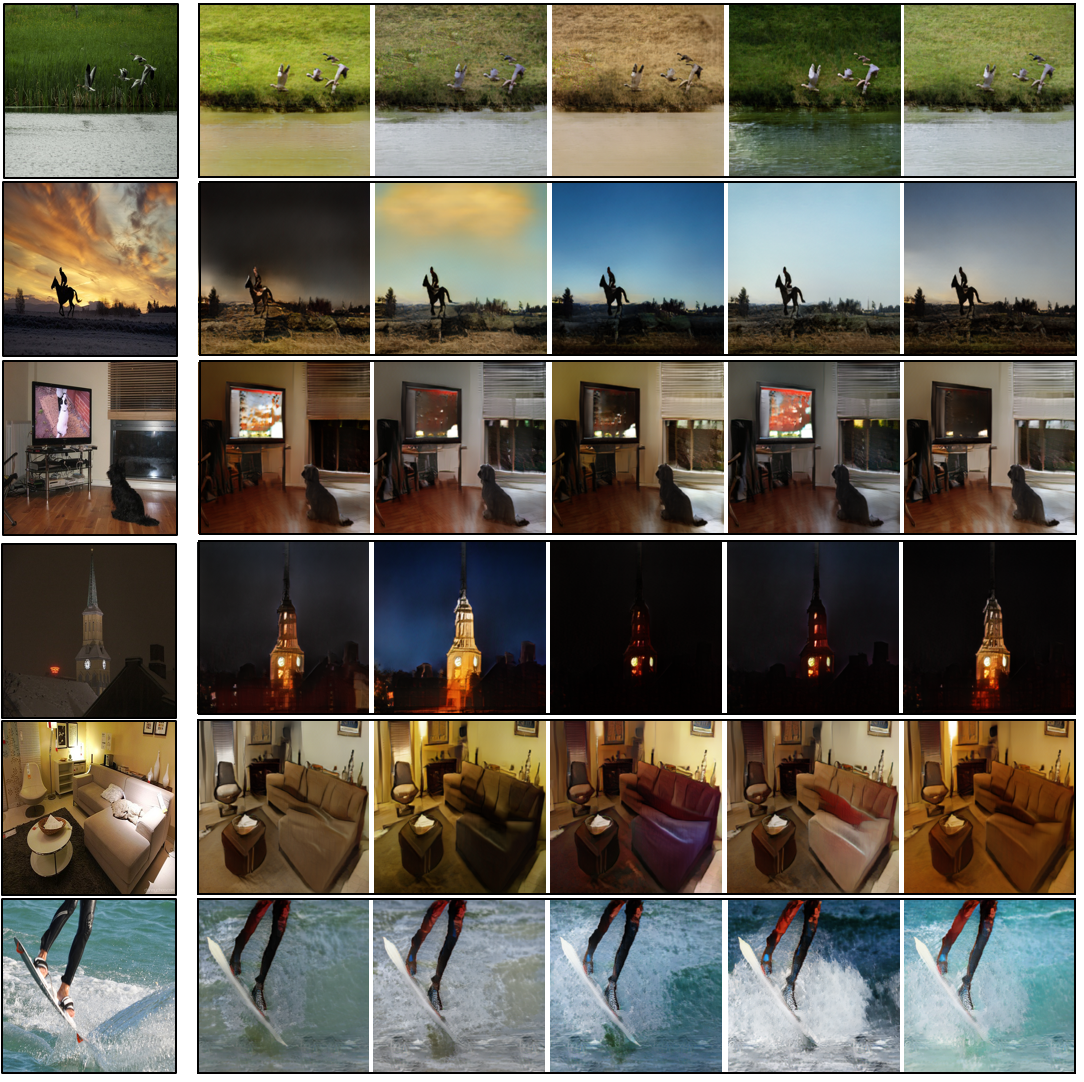}
    \vspace{-8pt}
    \caption{Example generated images from the COCO dataset: in each case, the left column is an original image, and the other columns show the repainted versions.}
    \label{fig:examples-coco-2}
\end{figure*}

\begin{figure*}
    \includegraphics[width=1.0\linewidth]{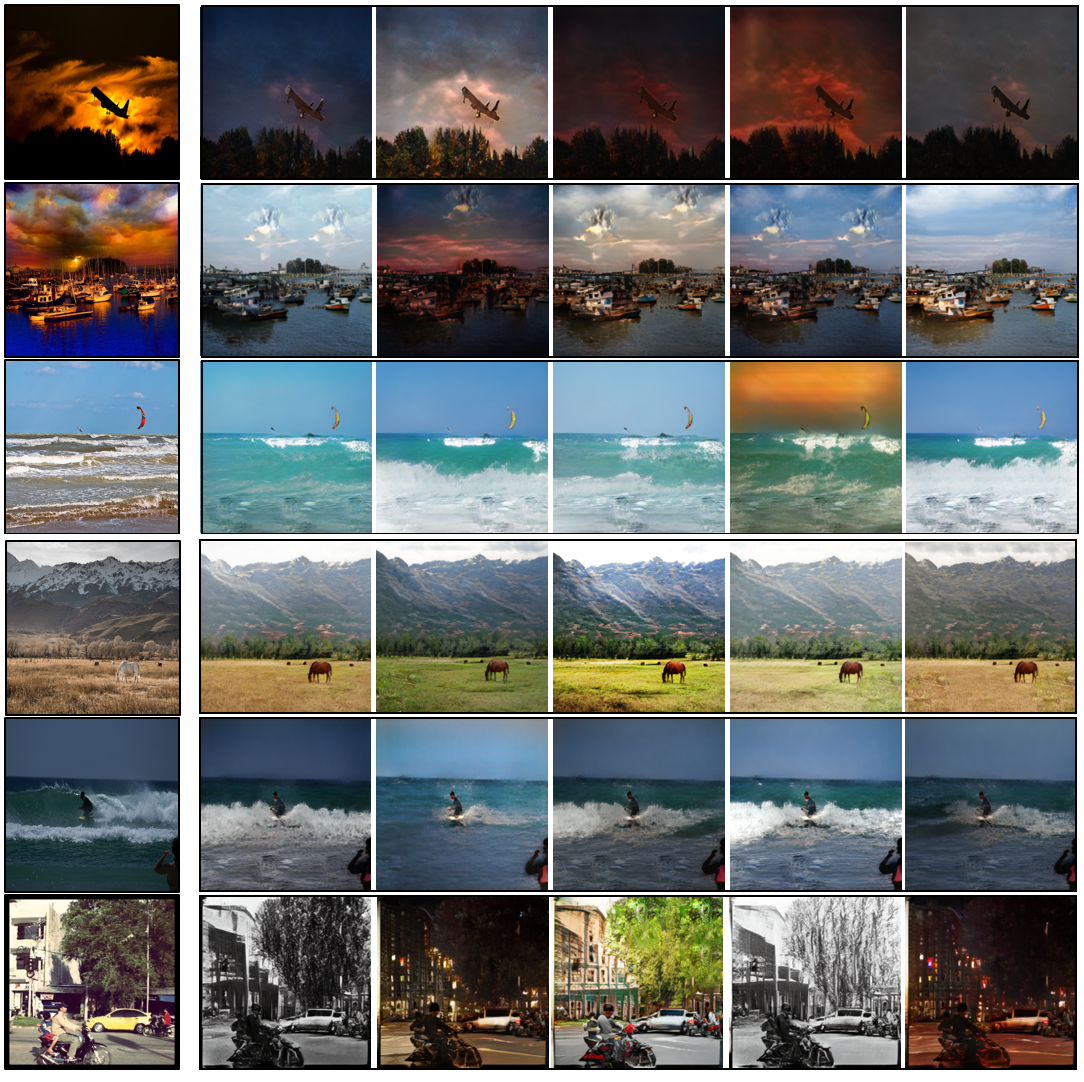}
    \vspace{-8pt}
    \caption{Example generated images from the COCO dataset: in each case, the left column is an original image, and the other columns show the repainted versions.}
    \label{fig:examples-coco-3}
\end{figure*}

\begin{figure*}
    \includegraphics[width=1.0\linewidth]{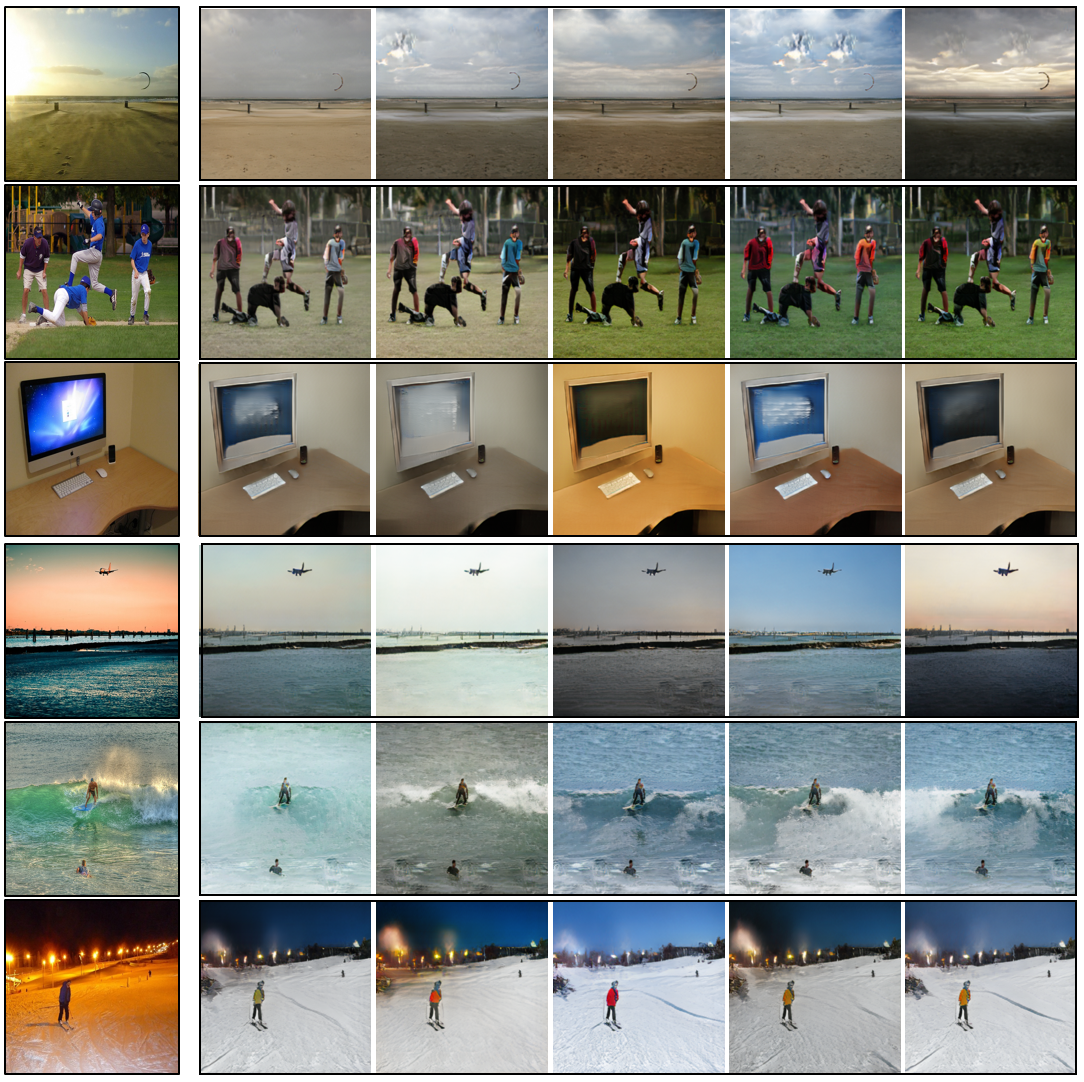}
    \vspace{-8pt}
    \caption{Example generated images from the COCO dataset: in each case, the left column is an original image, and the other columns show the repainted versions.}
    \label{fig:examples-coco-4}
\end{figure*}

\begin{figure*}
    \includegraphics[width=1.0\linewidth]{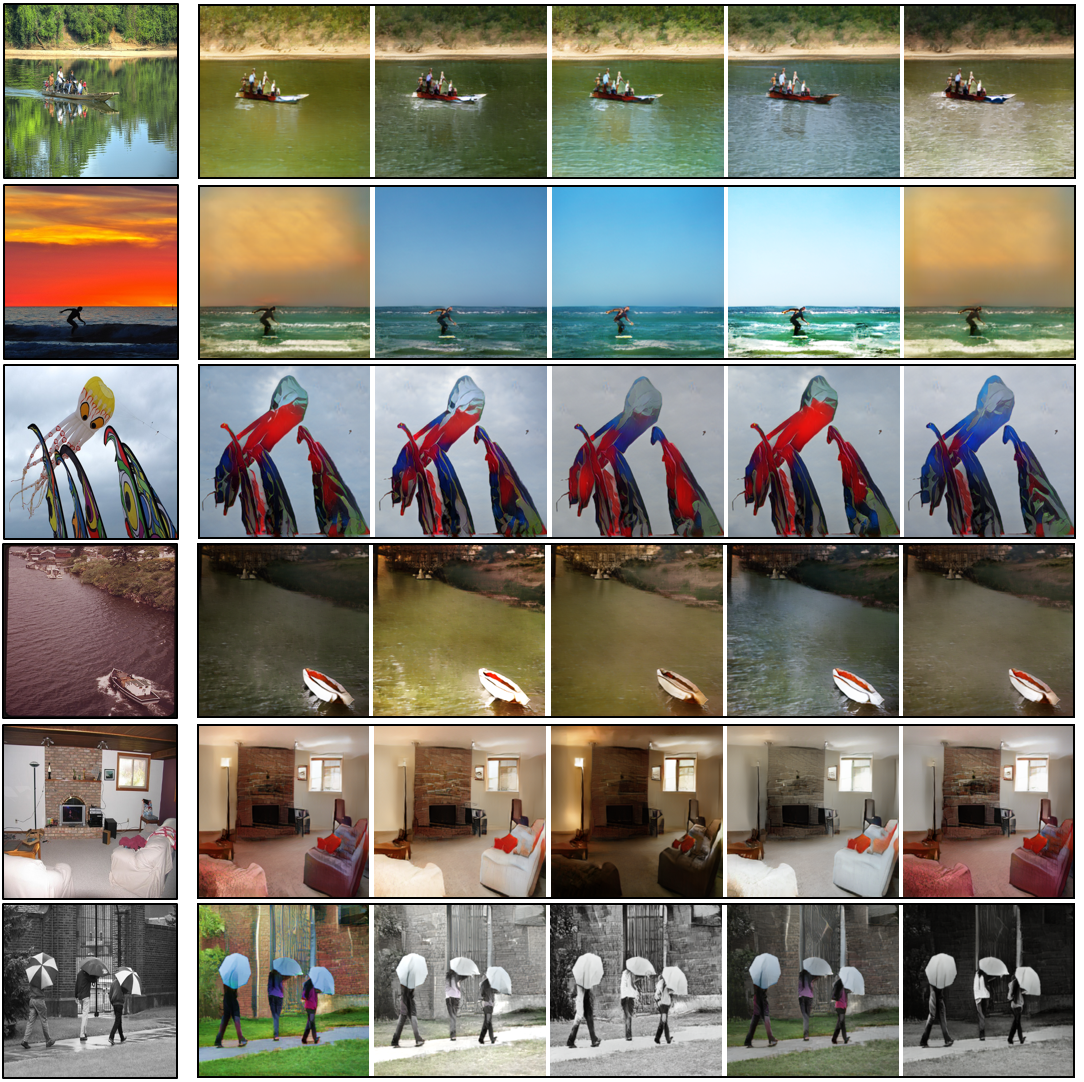}
    \vspace{-8pt}
    \caption{Example generated images from the COCO dataset: in each case, the left column is an original image, and the other columns show the repainted versions.}
    \label{fig:examples-coco-5}
\end{figure*}

\begin{figure*}
    \includegraphics[width=1.0\linewidth]{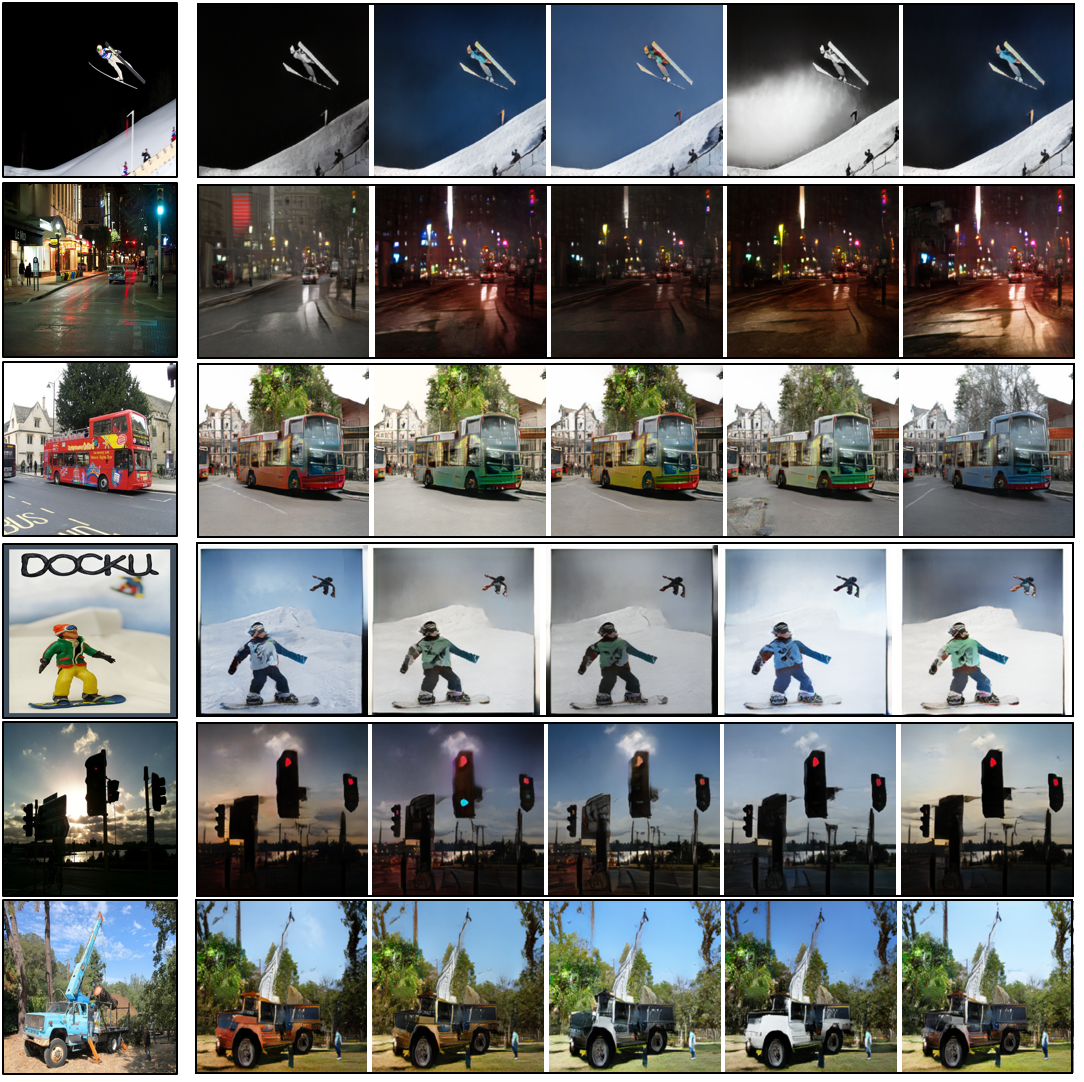}
    \vspace{-8pt}
    \caption{Example generated images from the COCO dataset: in each case, the left column is an original image, and the other columns show the repainted versions.}
    \label{fig:examples-coco-6}
\end{figure*}

\begin{figure*}
    \includegraphics[width=1.0\linewidth]{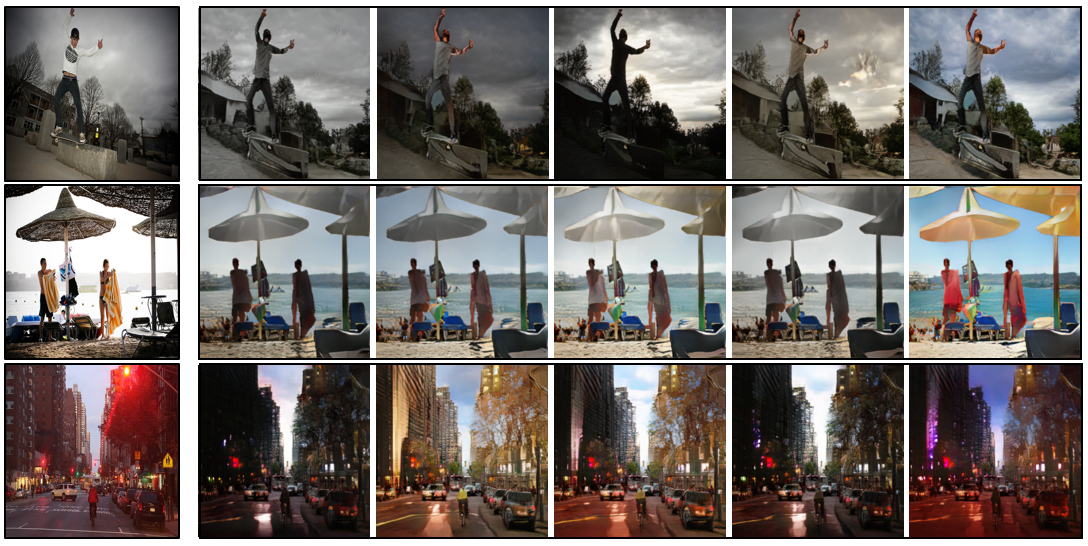}
    \vspace{-8pt}
    \caption{Example generated images from the COCO dataset: in each case, the left column is an original image, and the other columns show the repainted versions.}
    \label{fig:examples-coco-7}
\end{figure*}

\begin{figure*}
    \includegraphics[width=1.0\linewidth]{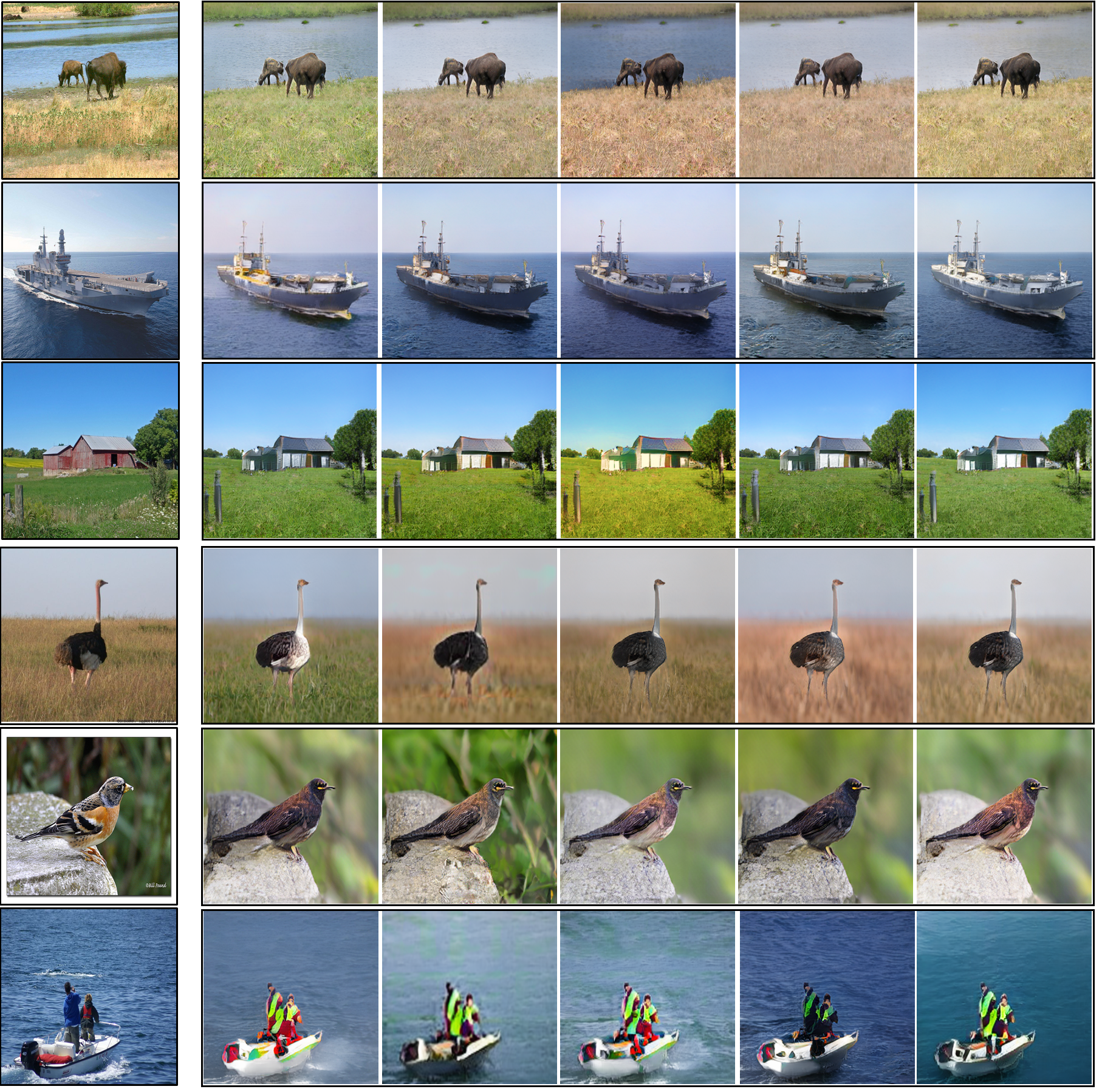}
    \vspace{-8pt}
    \caption{Example generated images from the ImageNet dataset: in each case, the left column is an original image, and the other columns show the repainted versions.}
    \label{fig:examples-imagenet-1}
\end{figure*}

\begin{figure*}
    \includegraphics[width=1.0\linewidth]{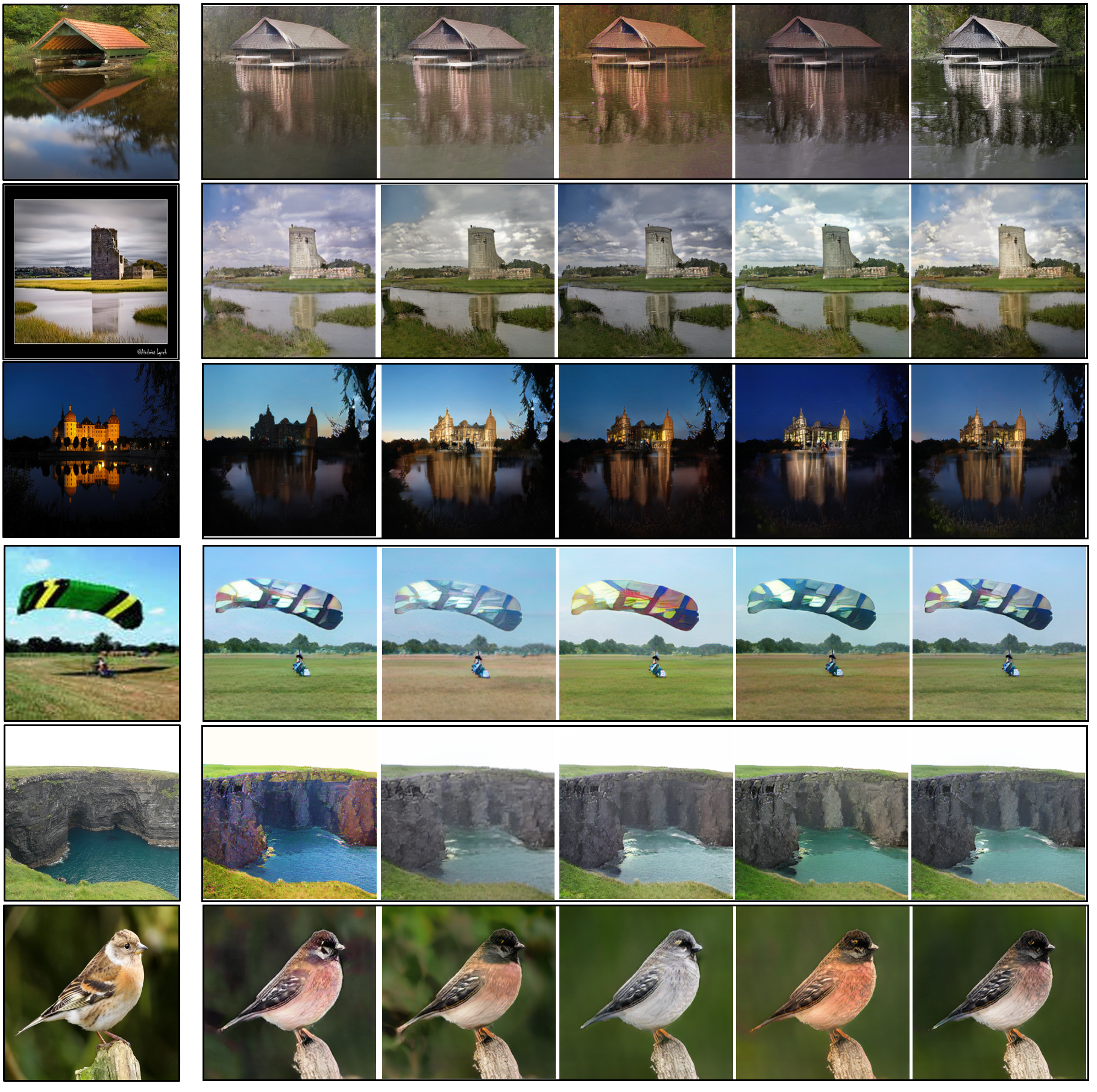}
    \vspace{-8pt}
    \caption{Example generated images from the ImageNet dataset: in each case, the left column is an original image, and the other columns show the repainted versions.}
    \label{fig:examples-imagenet-2}
\end{figure*}

\begin{figure*}
    \includegraphics[width=1.0\linewidth]{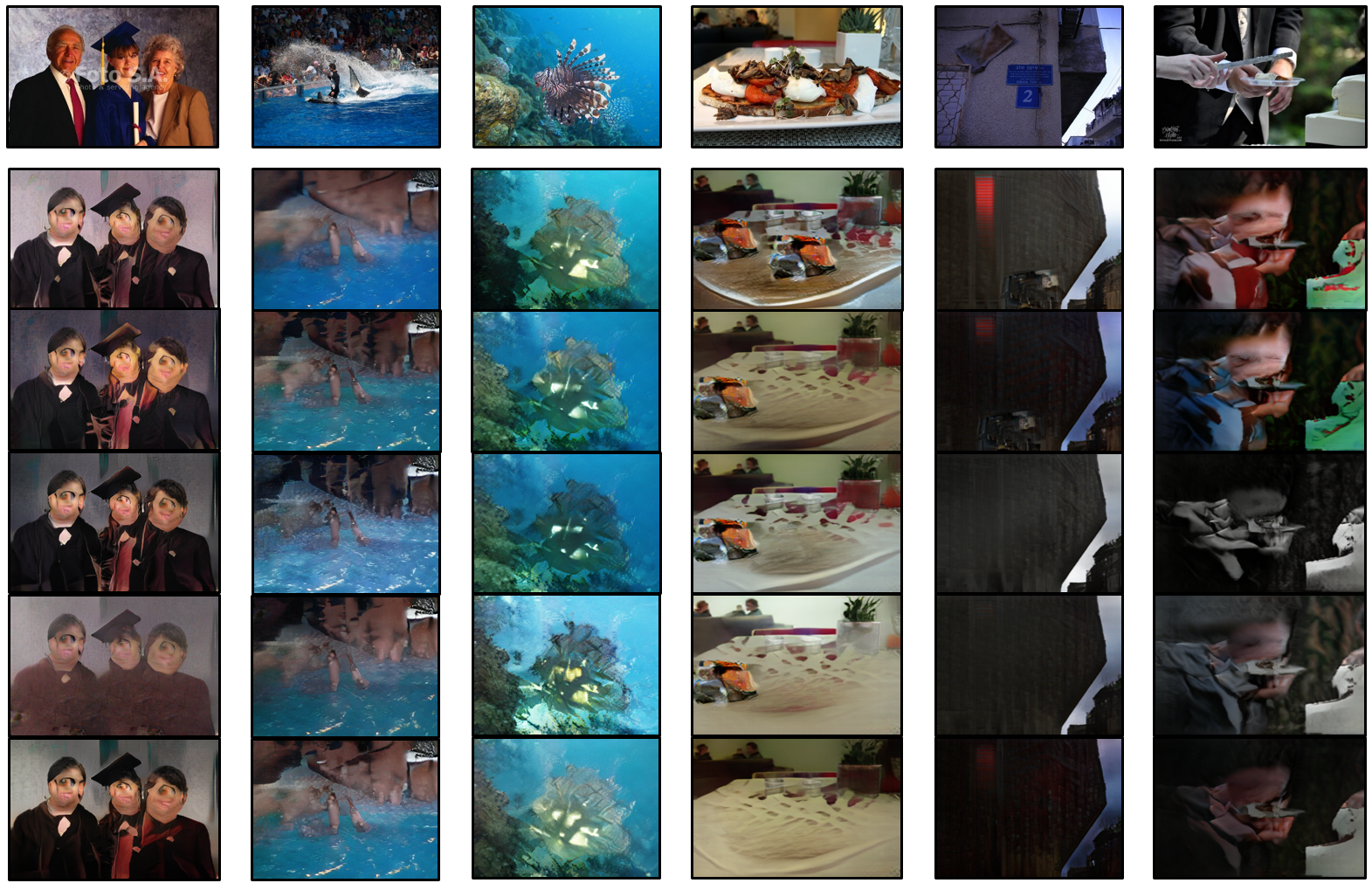}
    \vspace{-8pt}
    \caption{Examples of failure cases. In case of fine details such as facial features (in a natural diverse dataset like COCO or ImageNet) or in case of very rare objects, sometimes the algorithm fails to generate good-looking images. Nonetheless, the learning objective is to do well on the down-stream task, and not exactly only on image generation.}
    \label{fig:examples-failures}
\end{figure*}

Next, we provide a large set of additional visualizations. Figure \ref{fig:examples-zoom} shows patches of original images and several repainted versions. We observe from this figure the diversification of texture patterns of stripes on a zebra, parachute, and sea wave. Figure \ref{fig:examples-pairs-1} and \ref{fig:examples-pairs-2} illustrate additional pairs of original and repainted images. Figure \ref{fig:examples-coco-1},  \ref{fig:examples-coco-2}, \ref{fig:examples-coco-3}, \ref{fig:examples-coco-4}, \ref{fig:examples-coco-5}, \ref{fig:examples-coco-6}, and \ref{fig:examples-coco-7} demonstrate additional visualization for multiple instances generation from the COCO dataset. Similarly, Figure  \ref{fig:examples-imagenet-1} and  \ref{fig:examples-imagenet-2} show instance generation from the ImageNet dataset. In Figure \ref{fig:examples-failures}, we provide several examples of failure in generating good-looking images, and explain that it's not directly the main goal of the model to generate natural or good-looking images.



\end{document}